\documentclass{article} 
\usepackage{iclr2019_conference,times}

\usepackage{amsmath,amsfonts,bm}

\def\eqref#1{equation~\ref{#1}}

\def\1{\bm{1}}

\DeclareMathAlphabet{\mathsfit}{\encodingdefault}{\sfdefault}{m}{sl}
\SetMathAlphabet{\mathsfit}{bold}{\encodingdefault}{\sfdefault}{bx}{n}

\def\gA{{\mathcal{A}}}

\def\gM{{\mathcal{M}}}

\def\gS{{\mathcal{S}}}
\def\gT{{\mathcal{T}}}

\def\sP{{\mathbb{P}}}

\newcommand{\E}{\mathbb{E}}

\usepackage{hyperref}
\usepackage{url}
\usepackage{soul, color}
\usepackage{graphicx}
\usepackage{wrapfig}

\iclrfinalcopy 

\title{How Transferable are the Representations Learned by Deep Q Agents?}

\author{Jacob Tyo \& Zachary Lipton \\ 
Machine Learning Department\\
Carnegie Mellon University \\
Pittsburgh, PA 15213, USA \\
\texttt{jtyo@cs.cmu.edu, zlipton@cmu.edu} \\
}

\begin{document}

\maketitle

\begin{abstract}
In this paper, we consider the source of 
Deep Reinforcement Learning (DRL)'s sample complexity,
asking how much derives from the requirement 
of learning useful representations of environment states
and how much is due to the sample complexity of learning a policy.
While for DRL agents, the distinction between representation and policy may not be clear,
we seek new insight through a set of transfer learning experiments. 
In each experiment, we retain some fraction of layers 
trained on either the same game or a related game,
comparing the benefits of transfer learning to learning a policy from scratch.  
Interestingly, we find that benefits due to transfer 
are highly variable in general and non-symmetric across pairs of tasks.
Our experiments suggest that perhaps transfer from simpler environments 
can boost performance on more complex downstream tasks 
and that the requirements of learning a useful representation can range from 
negligible to the majority of the sample complexity, based on the environment.
Furthermore, we find that fine-tuning generally outperforms 
training with the transferred layers frozen,
confirming an insight first noted in the classification setting.
\end{abstract}

\section{Introduction}
\label{sec:intro}
Deep Reinforcement Learning (DRL) agents learn policies by selecting actions directly from raw perceptual data.
Despite numerous breakthroughs \citep{mnih2015human, alphastarblog, silver2016mastering}, 
DRL's prohibitive sample complexity 
limits its real world application.
The sample complexity of DRL may derive 
from many sources, including the requirement of learning useful state representations and the requirement of learning good policies given a suitable representation.  
This paper provides several simple experiments 
to provide new insights into this breakdown.



For a DRL agent, where precisely the ``representation'' ends 
and the ``policy'' begins is not clear.
In our experiments, we consider multiple interpretations 
of this divide by partitioning the network at various layers. 
To evaluate the extent to which representation learning 
contributes to the sample complexity of DRL, 
we execute a series of transfer learning experiments 
aimed to determine how quickly an agent can learn
given pre-learned representations (from either the same or a different game).
Our experiments proceed in the following manner:
1) Train a parent network until best reported performance is achieved. 
2) Transplant the first $k$ layers into a child network, 
re-initializing the remaining $l-k$ layers randomly. 
3) Train the child network, either fine-tuning the transplanted layers 
or keeping them frozen, following the methodology of \citet{yosinski2014transferable}.  

In particular, our experiments address the transferability of DQN representations 
among the three Atari games \emph{Berzerk}, \emph{Krull}, and \emph{River Raid}. 
We find the benefits of transferring representations from pretrained networks 
to be remarkably variable in general and non-symmetric across pairs of tasks. 
While we have not evaluated enough tasks to draw definitive conclusions, 
our preliminary results suggest that transfer from simple environments 
may improve performance on more complex tasks more than the reverse transfer.
Furthermore, our results suggest that the contribution of learning a useful representation to the overall sample complexity of the problem can range from negligible to the majority, based on the destination environment.  
Lastly, fine-tuning the transferred layers outperforms 
training with those layers frozen in general.

\section{Background and Experimental Setup}
\label{sec:methods}
Reinforcement Learning (RL) methods address the problem 
where an \emph{agent} must learn to act in an unknown \emph{environment}
to maximize a \emph{reward} signal.  
Initially, the environment provides a state $s_0$ to the agent, 
and then the agent selects an action $a_0$ based on the provided state.  
Thereafter, at each time step the environment provides an agent 
with a state $s_{t}$ and a reward $r_{t}$ each influenced by the agents previous action.
The agent must then select subsequent actions $a_t$, and so on, until the episode terminates.  
Formally, this interaction is described by a Markov Decision Process (MDP)
$\gM = \langle \gS, \gA, \gT, r, \gamma \rangle$, 
where $\gS$ is the set of states,
$\gA$ is the set of actions,
$\gT (s, a, s') = \sP(s_{t+1} = s' | s_t = s, a_t = a)$ is the transition function, 
$r(s, a) = \E[r_{t+1} | s_t = s, a_t = a]$ is the reward function, 
and $\gamma \in [0, 1]$ is a discount factor.  

We focus on Q-learning, an off-policy value-based RL algorithm.  
The value of each state is represented by $v^\pi(s) = \E_\pi[G_t(s_t) | s_t = s]$, and the value of each state-action pair is represented by $q^\pi(s, a) = \E_\pi[G_t(s_t) | s_t = s, a_t = a]$ where $G_t(s) = \sum_{i=t}^{T} \gamma^{i-t}r_t(s)$.  
In Q-learning, the agent estimates the state-action value function by
predicting the expected discounted return. 

Many interesting problems, including Atari games,
have a large state and action space, 
making tabular estimates of the Q-function intractable. 
In these cases, the Q-function can be approximated.
DRL denotes methods that approximate either the value function
(or, in other algorithms, the policy directly) by deep neural networks. 

In this paper, we build on a DRL Q-learning implementation called \emph{Rainbow}~\citep{hessel2018rainbow}: 
a 5-layer convolutional neural network based on~\citet{mnih2015human}
that incorporates double Q-learning~\citep{van2016deep}, 
prioritized replay~\citep{schaul2015prioritized}, 
dueling networks~\citep{wang2015dueling},
multi-step learning~\citep{sutton1988learning}, 
distributional RL~\citep{bellemare2017distributional}, 
and Noisy Nets~\citep{fortunato2017noisy}.  
Furthermore, Rainbow maintains two separate Q-networks: 
one with parameters $\theta$, and a second with parameters $\theta_\text{target}$ 
that are updated from $\theta$ every fixed number of iterations.  
In order to capture the game dynamics, a state is represented 
by a sequence (four in our case) of history frames.  


We tested the transferability of features learned 
by Rainbow agents on Atari games (i.e. environments).  
To make transfer learning experiments feasible within our resources,
we selected environments according to the speed 
that a Rainbow agents can reach high performance, 
requiring that the cardinality of the state and action spaces 
of each environment be equivalent 
and that two among the three environments were qualitatively similar (same genre of game).
We are interested in game similarity
to test a hypothesis that more similar games have more similar representations, and therefore agent transfer should be more effective between them. 
Environments with the same state and action space cardinality is required as it made agent transfer possible without modification.  
Pulling from the results of \citet{hessel2018rainbow}, we selected \textit{Berzerk}, \textit{Krull}, and \textit{River Raid} from the Arcade Learning Environment~\citep{bellemare2013arcade}.

Drawing inspiration from \citet{yosinski2014transferable}, our experiments proceeded as follows: 
\begin{enumerate}
    \item For all environments, train a parent network (a Rainbow agent) until best-reported performance is reached.
    \item For every permutation of environment pairs, transplant the first $k$ layers of the parent network into a child network (also a Rainbow agent), then reinitialize the remaining $l-k$ layers randomly (where $l$ is the total number of layers in the network and $k \in \{ 2, 4 \}$).
    \item Train the child network, either fine-tuning or freezing the transplanted portion of the network (we explore both settings).  
\end{enumerate}
With $3$ runs per pair of environments, setting of $k$, and each choice among \{freezing, fine-tuning\}, 
we ran a total of $111$ trials taking over $35$ days on the available resources.

\section{Experimental Results and Discussion}
\label{sec:experiments}
In this section, we analyze the performance 
and learned representations of the child agents, 
where one child exists for every environment pair, $k$-value, 
and choice among freezing/fine-tuning.  
For brevity, we will denote child agents with 4 layers transferred from a parent trained on environment1 then frozen as child4-frozen-environment1.
We will refer to the parents as baselines with respect
to their performance on the environment that they were trained.

%
%
We evaluate each network separated at two places, transferring $2$, and transferring $4$ (out of $5$) layers 
and compare freezing vs fine-tuning over the transplanted layers.
For each respective experiment, we refer 
to the output of the last layer 
of the transplanted portion of the network 
as the ``representation.'' 






\begin{figure}[t]
    \centering
    \includegraphics[width=\textwidth]{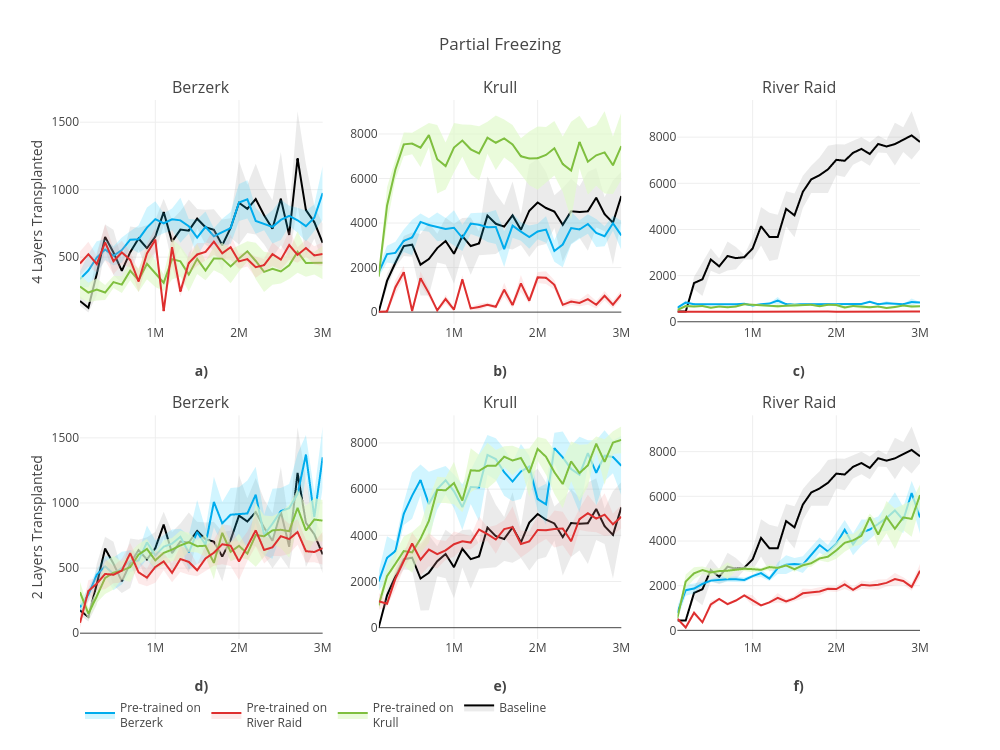}
    \caption{Child agent (blue, red, and green) performance over iterations of training when the transplanted layers are frozen. Compare against the parent network trained from scratch (black).}
    \label{fg:partialFreezing}
\end{figure}

Figures~\ref{fg:partialFreezing}a,~\ref{fg:partialFreezing}b, and \ref{fg:partialFreezing}c
demonstrate the performance of all environment pairs
when $4$ layers are transplanted and then frozen during subsequent training.  
Because only the output layer is not frozen,
these plots show the performance of a linear policy.
trained on the transplanted representation.  
\citet{levine2017shallow} deem representations 
that can be used with a linear policy as a ``good'' representation, 
thus our analysis starts with these corresponding experiments.  


Figure~\ref{fg:partialFreezing}a
presents the performance of the agents on Berzerk
for runs with $4$ transplanted layers,
all of which are frozen during the subsequent training.  
The child4-frozen-Berzerk agent (blue) does not out perform the baseline (either by final performance or training speed), 
which we find surprising, since four of five layers 
of this agent 
were transferred from a high-performing parent trained on the same environment, leaving only a linear policy to be relearned. 
We might deduce that for this game, the entire difficulty of DRL 
can be attributed simply to the ``RL'' since starting off 
with the representations upon which a strong linear policy can be learnt 
seems to confer no benefit. 
%
Notably,  
the agents transferred from foreign games (child4-frozen-Krull and child4-frozen-RiverRaid) have worse final performance than the baseline.  
As a linear policy could not be learnt from a known good representation, we would not expect any other representation to perform better.  
Figure~\ref{fg:partialFreezing}d shows these trends continuing when $2$ layers are transplanted.
The difference in performance between the children and the baseline is negligible, 
and therefore indicating that the difficulty does lie within the RL problem. 
Even with more model complexity allowed to the policy and different representation, 
no benefits are seen.

However, as shown in Figure~\ref{fg:partialFreezing}b, 
the transfer from Krull-to-Krull results in faster training and higher final performance than the baseline, 
perhaps suggesting that for this game, 
representation learning is a more significant part of the challenge.  
Note that with only two layers transferred, 
the representations pretrained on both Krull and Berzerk, 
outperform the baseline as shown in Figure~\ref{fg:partialFreezing}e.
This reflects our intuition that these games are similar,
while the representations pretrained on River Raid confer no benefit. 


\begin{figure}[t]
    \centering
    \includegraphics[width=\textwidth]{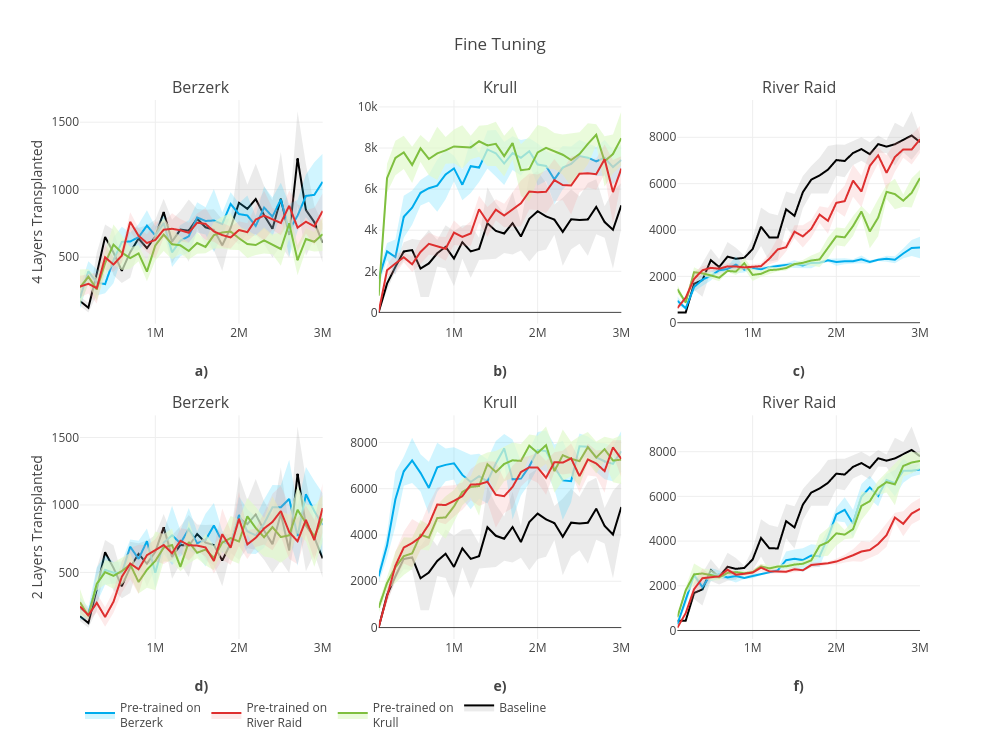}
    \caption{Child agent (blue, red, and green) performance over iterations of training when the transplanted layers are fine-tuned. Compare against the parent network trained from scratch (black).} 
    \label{fg:fineTuning}
\end{figure}


Transferring and freezing $4$ of $5$ layers from the parent to child agent on Berzerk and Krull showed negligible and positive changes in performance, respectively.  
However, River Raid paints a different picture.  
As shown in Figure~\ref{fg:partialFreezing}c, 
no agent is able to learn a useful policy.  
This is especially surprising because we know that if nothing else, 
the transfer from the parent trained on River Raid itself can replicate the baseline results by learning a linear policy.  
Similarly, we see in Figure~\ref{fg:partialFreezing}f that the agent with transplanted layers trained on the \emph{same} environment performs the worst.  
Again this is surprising as, theoretically, the baseline could be replicated. 
Interestingly, this may indicate that a representation learned from a more simple environment may be helpful in solving more complex environments.  
The converse does not appear to hold from our experiments,
but a deeper analysis on more tasks would be required to confirm these intuitions.  




In general, we see that when transplanted layers are frozen, 
there is a trade-off between negative transfer and training speed.  
As the number of unfrozen layers increases, 
the effect of negative transfer decreases 
but the training speed slows slightly.  
Furthermore, Figure~\ref{fg:partialFreezing}
shows that transfer between environments is not symmetric.  
In other words, if a representation $x$ is learned on environment $X$ 
and is shown to perform well on environment $Y$, 
then this \emph{does not} imply that a representation $y$ 
learned on environment $Y$ will be effective on $X$.  
This relation continues to hold, 
even when fine-tuning over the transplanted layers. 
However, when fine-tuning over the transplanted layers 
we do not see the same level of negative transfer.

Figures~\ref{fg:fineTuning}a, \ref{fg:fineTuning}b, and \ref{fg:fineTuning}c 
show the performance of all environment pairs 
when four layers are transplanted from the parents to the children and then frozen, 
whereas Figures \ref{fg:fineTuning}d, \ref{fg:fineTuning}e, and \ref{fg:fineTuning}f 
show the performance when only two layers are transplanted and frozen.   
Intuitively, these experiments which allow 
the transplanted representations to change during training, 
examine a more flexible notion of their utility (as initializations only).
For these experiments, in all environments except for River Raid, 
negative transfer is no longer an issue.  
In River Raid, negative transfer is seen in $2$ of $6$ children, 
vs in $6$ of $6$ children when the transferred layers are frozen.

The performance of all agents on Berzerk 
with $4$ and $2$ layers transplanted and fine-tuned respectively, 
shown in Figures~\ref{fg:fineTuning}a and \ref{fg:fineTuning}d, 
are similar to the performance seen for the same game under frozen transplanted layers.
There is little difference between each of the agents, 
regardless of the environment on which the parent was trained.  
This further supports the earlier conclusions
that the difficulty of this game is due entirely 
to learning the policy as no transferred representation improves performance. 



Transplanting and fine-tuning layers results in a large decrease in training time, 
and a large increase in final performance on the Krull environment.  
This holds for both transplanting and fine-tuning 
over $4$ layers as shown in Figure~\ref{fg:fineTuning}b, 
and over $2$ layers as shown in Figure~\ref{fg:fineTuning}e.  
As expected, the transplanted layers from the parent trained on Krull 
performed the best when $4$ layers were transplanted.
But interestingly when only $2$ layers were transplanted, 
the transplanted layers from the parent trained on Berzerk were more effective.  
Overall, this supports our earlier conclusion of 
the importance of representations for Krull---in all cases, 
transfer is preferred to random initialization.




In  Figures \ref{fg:partialFreezing} and \ref{fg:fineTuning}, we see the general trend that the transfer of pretrained layers 
has negligible effect on Berzerk, 
positive effects on Krull, and negative effects on River Raid.  
Figures~\ref{fg:fineTuning}c and \ref{fg:fineTuning}f show the performance of all agents with $4$ and $2$ layers transplanted and fine-tune respectively.  
Interestingly, the trend observed in Krull, 
where the best performing child was the one with features transferred from a parent trained on the same environment is the best performer when $4$ layers are transferred, 
yet it is not the best performer when only $2$ layers are transferred, 
is also observed with River Raid.  However, the baseline trains faster and reaches better final performance than all of the children on River Raid.

\section{Conclusions}
\label{sec:Conclusions}
This paper presents an empirical evaluation of $111$ transfer learning experiments on agents trained on the Atari 2600 games Berzerk, Krull, and River Raid.  We compare the effect of transplanting initial layers of a pretrained network into a child network while either freezing or fine-tuning over the transplanted layers.  
Surprisingly, the benefits of transferring portions of pretrained networks are highly variable and non-symmetric across tasks. 
Furthermore, the requirements of learning a useful representation can range from nothing to the majority of the sample complexity based on the destination environment. 
We present analyses for why each task transfer occurs as shown, and give intuition for understanding representations and policies in DQNs.  
We show that, in general, fine tuning is better than freezing portions of networks, as performance gains can still be expected with less likelihood of negative transfer.  


\citet{zahavy2016graying} have shown that DQNs find hierarchical abstractions automatically.  
Our work suggests that the similarity of the high level task abstraction may be a good metric to determine the transferabitliy of DQN agents on.  
Future work includes heavier analysis on these experiments to determine how agents pretrained on different environments ``focus'' differently, how their representations differ, and how to numerically quantify the contribution of representation learning to the overall sample complexity.  
Furthermore, this methodology can pass insight to questions about the benefit of unsupervised reinforcement learning in pre-training, 
e.g. techniques based on intrinsic motivation ~\citep{chentanez2005intrinsically}.
To the extent that intrinsic motivation serves to learn representations suitable for fine-tuning to a given reward signal, 
quantifying just how much representation learning is the bottleneck to learning in the first place can provide insight in assessing its potential. 



\clearpage

\bibliography{references.bib}

\begin{thebibliography}{23}
\providecommand{\natexlab}[1]{#1}
\providecommand{\url}[1]{\texttt{#1}}
\expandafter\ifx\csname urlstyle\endcsname\relax
  \providecommand{\doi}[1]{doi: #1}\else
  \providecommand{\doi}{doi: \begingroup \urlstyle{rm}\Url}\fi

\bibitem[Asadi \& Huber(2007)Asadi and Huber]{asadi2007effective}
Mehran Asadi and Manfred Huber.
\newblock Effective control knowledge transfer through learning skill and
  representation hierarchies.
\newblock In \emph{IJCAI}, volume~7, pp.\  2054--2059, 2007.

\bibitem[Bellemare et~al.(2013)Bellemare, Naddaf, Veness, and
  Bowling]{bellemare2013arcade}
Marc~G Bellemare, Yavar Naddaf, Joel Veness, and Michael Bowling.
\newblock The arcade learning environment: An evaluation platform for general
  agents.
\newblock \emph{Journal of Artificial Intelligence Research}, 47:\penalty0
  253--279, 2013.

\bibitem[Bellemare et~al.(2017)Bellemare, Dabney, and
  Munos]{bellemare2017distributional}
Marc~G Bellemare, Will Dabney, and R{\'e}mi Munos.
\newblock A distributional perspective on reinforcement learning.
\newblock In \emph{Proceedings of the 34th International Conference on Machine
  Learning-Volume 70}, pp.\  449--458. JMLR. org, 2017.

\bibitem[Calandriello et~al.(2014)Calandriello, Lazaric, and
  Restelli]{calandriello2014sparse}
Daniele Calandriello, Alessandro Lazaric, and Marcello Restelli.
\newblock Sparse multi-task reinforcement learning.
\newblock In \emph{Advances in Neural Information Processing Systems}, pp.\
  819--827, 2014.

\bibitem[Chentanez et~al.(2005)Chentanez, Barto, and
  Singh]{chentanez2005intrinsically}
Nuttapong Chentanez, Andrew~G Barto, and Satinder~P Singh.
\newblock Intrinsically motivated reinforcement learning.
\newblock In \emph{Advances in neural information processing systems}, pp.\
  1281--1288, 2005.

\bibitem[Fortunato et~al.(2017)Fortunato, Azar, Piot, Menick, Osband, Graves,
  Mnih, Munos, Hassabis, Pietquin, et~al.]{fortunato2017noisy}
Meire Fortunato, Mohammad~Gheshlaghi Azar, Bilal Piot, Jacob Menick, Ian
  Osband, Alex Graves, Vlad Mnih, Remi Munos, Demis Hassabis, Olivier Pietquin,
  et~al.
\newblock Noisy networks for exploration.
\newblock \emph{arXiv preprint arXiv:1706.10295}, 2017.

\bibitem[Foster \& Dayan(2002)Foster and Dayan]{foster2002structure}
David Foster and Peter Dayan.
\newblock Structure in the space of value functions.
\newblock \emph{Machine Learning}, 49\penalty0 (2-3):\penalty0 325--346, 2002.

\bibitem[Hessel et~al.(2018)Hessel, Modayil, Van~Hasselt, Schaul, Ostrovski,
  Dabney, Horgan, Piot, Azar, and Silver]{hessel2018rainbow}
Matteo Hessel, Joseph Modayil, Hado Van~Hasselt, Tom Schaul, Georg Ostrovski,
  Will Dabney, Dan Horgan, Bilal Piot, Mohammad Azar, and David Silver.
\newblock Rainbow: Combining improvements in deep reinforcement learning.
\newblock In \emph{Thirty-Second AAAI Conference on Artificial Intelligence},
  2018.

\bibitem[Lazaric(2008)]{lazaric2008knowledge}
Alessandro Lazaric.
\newblock \emph{Knowledge transfer in reinforcement learning}.
\newblock PhD thesis, PhD thesis, Politecnico di Milano, 2008.

\bibitem[Levine et~al.(2017)Levine, Zahavy, Mankowitz, Tamar, and
  Mannor]{levine2017shallow}
Nir Levine, Tom Zahavy, Daniel~J Mankowitz, Aviv Tamar, and Shie Mannor.
\newblock Shallow updates for deep reinforcement learning.
\newblock In \emph{Advances in Neural Information Processing Systems}, pp.\
  3135--3145, 2017.

\bibitem[Mnih et~al.(2015)Mnih, Kavukcuoglu, Silver, Rusu, Veness, Bellemare,
  Graves, Riedmiller, Fidjeland, Ostrovski, et~al.]{mnih2015human}
Volodymyr Mnih, Koray Kavukcuoglu, David Silver, Andrei~A Rusu, Joel Veness,
  Marc~G Bellemare, Alex Graves, Martin Riedmiller, Andreas~K Fidjeland, Georg
  Ostrovski, et~al.
\newblock Human-level control through deep reinforcement learning.
\newblock \emph{Nature}, 518\penalty0 (7540):\penalty0 529, 2015.

\bibitem[Rusu et~al.(2016)Rusu, Rabinowitz, Desjardins, Soyer, Kirkpatrick,
  Kavukcuoglu, Pascanu, and Hadsell]{rusu2016progressive}
Andrei~A Rusu, Neil~C Rabinowitz, Guillaume Desjardins, Hubert Soyer, James
  Kirkpatrick, Koray Kavukcuoglu, Razvan Pascanu, and Raia Hadsell.
\newblock Progressive neural networks.
\newblock \emph{arXiv preprint arXiv:1606.04671}, 2016.

\bibitem[Schaul et~al.(2015)Schaul, Quan, Antonoglou, and
  Silver]{schaul2015prioritized}
Tom Schaul, John Quan, Ioannis Antonoglou, and David Silver.
\newblock Prioritized experience replay.
\newblock \emph{arXiv preprint arXiv:1511.05952}, 2015.

\bibitem[Silver et~al.(2016)Silver, Huang, Maddison, Guez, Sifre, Van
  Den~Driessche, Schrittwieser, Antonoglou, Panneershelvam, Lanctot,
  et~al.]{silver2016mastering}
David Silver, Aja Huang, Chris~J Maddison, Arthur Guez, Laurent Sifre, George
  Van Den~Driessche, Julian Schrittwieser, Ioannis Antonoglou, Veda
  Panneershelvam, Marc Lanctot, et~al.
\newblock Mastering the game of go with deep neural networks and tree search.
\newblock \emph{nature}, 529\penalty0 (7587):\penalty0 484, 2016.

\bibitem[Singh(1992)]{singh1992transfer}
Satinder~Pal Singh.
\newblock Transfer of learning by composing solutions of elemental sequential
  tasks.
\newblock \emph{Machine Learning}, 8\penalty0 (3-4):\penalty0 323--339, 1992.

\bibitem[Sutton(1988)]{sutton1988learning}
Richard~S Sutton.
\newblock Learning to predict by the methods of temporal differences.
\newblock \emph{Machine learning}, 3\penalty0 (1):\penalty0 9--44, 1988.

\bibitem[Taylor \& Stone(2009)Taylor and Stone]{taylor2009transfer}
Matthew~E Taylor and Peter Stone.
\newblock Transfer learning for reinforcement learning domains: A survey.
\newblock \emph{Journal of Machine Learning Research}, 10\penalty0
  (Jul):\penalty0 1633--1685, 2009.

\bibitem[Van~Hasselt et~al.(2016)Van~Hasselt, Guez, and Silver]{van2016deep}
Hado Van~Hasselt, Arthur Guez, and David Silver.
\newblock Deep reinforcement learning with double q-learning.
\newblock In \emph{Thirtieth AAAI Conference on Artificial Intelligence}, 2016.

\bibitem[Vinyals et~al.(2019)Vinyals, Babuschkin, Chung, Mathieu, Jaderberg,
  Czarnecki, Dudzik, Huang, Georgiev, Powell, Ewalds, Horgan, Kroiss,
  Danihelka, Agapiou, Oh, Dalibard, Choi, Sifre, Sulsky, Vezhnevets, Molloy,
  Cai, Budden, Paine, Gulcehre, Wang, Pfaff, Pohlen, Wu, Yogatama, Cohen,
  McKinney, Smith, Schaul, Lillicrap, Apps, Kavukcuoglu, Hassabis, and
  Silver]{alphastarblog}
Oriol Vinyals, Igor Babuschkin, Junyoung Chung, Michael Mathieu, Max Jaderberg,
  Wojciech~M. Czarnecki, Andrew Dudzik, Aja Huang, Petko Georgiev, Richard
  Powell, Timo Ewalds, Dan Horgan, Manuel Kroiss, Ivo Danihelka, John Agapiou,
  Junhyuk Oh, Valentin Dalibard, David Choi, Laurent Sifre, Yury Sulsky, Sasha
  Vezhnevets, James Molloy, Trevor Cai, David Budden, Tom Paine, Caglar
  Gulcehre, Ziyu Wang, Tobias Pfaff, Toby Pohlen, Yuhuai Wu, Dani Yogatama,
  Julia Cohen, Katrina McKinney, Oliver Smith, Tom Schaul, Timothy Lillicrap,
  Chris Apps, Koray Kavukcuoglu, Demis Hassabis, and David Silver.
\newblock {AlphaStar: Mastering the Real-Time Strategy Game StarCraft II}.
\newblock
  \url{https://deepmind.com/blog/alphastar-mastering-real-time-strategy-game-starcraft-ii/},
  2019.

\bibitem[Walsh et~al.(2006)Walsh, Li, and Littman]{walsh2006transferring}
Thomas~J Walsh, Lihong Li, and Michael~L Littman.
\newblock Transferring state abstractions between mdps.
\newblock In \emph{ICML Workshop on Structural Knowledge Transfer for Machine
  Learning}, 2006.

\bibitem[Wang et~al.(2015)Wang, Schaul, Hessel, Van~Hasselt, Lanctot, and
  De~Freitas]{wang2015dueling}
Ziyu Wang, Tom Schaul, Matteo Hessel, Hado Van~Hasselt, Marc Lanctot, and Nando
  De~Freitas.
\newblock Dueling network architectures for deep reinforcement learning.
\newblock \emph{arXiv preprint arXiv:1511.06581}, 2015.

\bibitem[Yosinski et~al.(2014)Yosinski, Clune, Bengio, and
  Lipson]{yosinski2014transferable}
Jason Yosinski, Jeff Clune, Yoshua Bengio, and Hod Lipson.
\newblock How transferable are features in deep neural networks?
\newblock In \emph{Advances in neural information processing systems}, pp.\
  3320--3328, 2014.

\bibitem[Zahavy et~al.(2016)Zahavy, Ben-Zrihem, and Mannor]{zahavy2016graying}
Tom Zahavy, Nir Ben-Zrihem, and Shie Mannor.
\newblock Graying the black box: Understanding dqns.
\newblock In \emph{International Conference on Machine Learning}, pp.\
  1899--1908, 2016.

\end{thebibliography}
\bibliographystyle{iclr2019_conference}

\clearpage
\begin{appendix}

\section{Related Work}
\label{sec:related}
\citet{hessel2018rainbow} present Rainbow, the DQN used in this paper, incorporating Double Q-learning~\citep{van2016deep}, prioritized replay~\citep{schaul2015prioritized}, dueling networks~\citep{wang2015dueling}, multi-step learning~\citep{sutton1988learning}, distributional reinforcement learning~\citep{bellemare2017distributional}, and noisy networks~\cite{fortunato2017noisy}.  

\citet{zahavy2016graying} analyze Deep Q Networks (DQNs) by observing the activation's of the model's last layer and saliency maps.  
They show that DQNs learn temporal abstractions, such as hierarchical state aggregation and options, automatically.  
\citet{levine2017shallow} show that the last layer in a deep architecture can be seen as a linear representation, and thus
can be learned using standard shallow reinforcement learning algorithms.  
They then show that this hybrid approach improves performance on the Atari benchmark.  

\citet{yosinski2014transferable} present a large scale study of feature transferabitlity in deep neural networks.  
They show that transferability is negatively affected by the specialization of higher layer neurons to their original task.  
Furthermore, optimization difficulties can arise when co-adapted neurons are split during transfer, and freezing vs fine-tuning over transferred layers are compared.  
The authors show that transferring features, even if from a very different task, can improve generalization performance even after substantial fine-tuning on a new task.  
Lastly, a relation between the effectiveness of transfer and the distance between tasks is presented, but even in the worst case is shown to be better than random.

\citet{taylor2009transfer} provides a survey of transfer learning techniques in reinforcement learning.  
Here we will focus on the tasks that allow variation in the reward function, as we assume the state spaces are of the same cardinality, and the action spaces are equivalent in this paper.  
\citet{singh1992transfer} and \cite{foster2002structure} learn multiple tasks by assuming that each goal (or composite) task is composed of several elemental tasks, and then learning a set of elemental tasks that can be composed to solve each task of interest.  

Solving multiple MDPs has also been approached from the representation perspective, specifically with the goal of developing a shared representation that can then be used to solve all tasks.  
The approach proposed by \citet{asadi2007effective} focuses on learning a more efficient state-space representation of the problem that will transfer between multiple tasks, and then learning options on the new representation. 
\citet{walsh2006transferring} use a similar approach, but rely on the learned state abstraction techniques to transfer between tasks.  
Another approach similar to state abstractions is to compare observations ($\langle s, a, r, s' \rangle)$ tuples from previous tasks to new tasks, then select the best action from the most similar previously experienced observation.  
\citet{lazaric2008knowledge} uses this approach in an attempt to generalize experiences from learned to novel tasks, and then \citet{calandriello2014sparse} extend this approach to include sparse representations.

\citet{rusu2016progressive} introduce progressive neural networks, which is a novel model architecture which retains a pool of pretrained models throughout training, and learns lateral connections from these to extract useful features for new tasks.  
This architecture leverages transfer learning while avoiding catastrophic forgetting and allowing for better incorporation of prior knowledge vs the traditional method of initialization.

\end{appendix}

\end{document}